\documentclass[journal]{IEEEtran}
\ifCLASSINFOpdf
 \else
  \fi

\def \R {\mathbb R}

\usepackage{multirow}
\usepackage{tabls}
\usepackage{wrapfig}

\usepackage{amsmath,amssymb} 
\usepackage{amsmath,amssymb}
 \usepackage[ruled,vlined,linesnumbered,resetcount]{algorithm2e}
 \usepackage[linesnumbered]{algorithm2e}
\usepackage{algorithmic}
\usepackage{caption}
\usepackage{subfigure}
\renewcommand{\algorithmicrequire}{ \textbf{Input:}} 
\renewcommand{\algorithmicensure}{ \textbf{Output:}} 
\usepackage[font=small,labelfont=bf]{caption}
   \usepackage{graphicx}
 \usepackage{booktabs}

\usepackage{epstopdf}
\usepackage{graphicx}

\begin{document}

\title{Sparse Representation-based Open Set Recognition}

\author{He Zhang,~\IEEEmembership{Student Member,~IEEE} and 
        Vishal M. Patel,~\IEEEmembership{Senior Member,~IEEE}
\thanks{This work has been accepted by T-PAMI.\quad He Zhang is with the department of Electrical and Computer Engineering at Rutgers University, 
   Piscataway, NJ USA. email: he.zhang92@rutgers.edu. }
\thanks{Vishal M. Patel is with the department of Electrical and Computer Engineering at Rutgers University, 
   Piscataway, NJ USA. email: vishal.m.patel@rutgers.edu.}
}

\markboth{IEEE TRANSACTIONS ON PATTERN ANALYSIS AND MACHINE INTELLIGENCE,Vol.~X, No.~X, ~2016}%
{Shell \MakeLowercase{\textit{et al.}}: Bare Demo of IEEEtran.cls for IEEE Journsssals}

\maketitle
\begin{abstract}
We propose a generalized Sparse Representation-based Classification (SRC) algorithm for open set recognition where not all classes presented during testing are known during training.  The SRC algorithm uses class reconstruction errors for classification.  As most of the discriminative information for open set recognition is hidden in the tail part of the matched and sum of non-matched reconstruction error distributions, we model the tail of those two error distributions using the statistical Extreme Value Theory (EVT).  Then we simplify the open set recognition problem into a set of hypothesis testing problems.   The confidence scores corresponding to the tail distributions of a novel test sample are then fused to determine its identity.  The effectiveness of the proposed method is demonstrated using four publicly available image and object classification datasets and it is shown that this method can perform significantly better than many competitive open set recognition algorithms. Code is public available: \emph{https://github.com/hezhangsprinter/SROSR}
\end{abstract}


\IEEEpeerreviewmaketitle

\section{Introduction}
In recent years, sparse representation-based techniques have drawn much interest in computer vision and image processing fields \cite{SR_IEEE_proc}, \cite{IEEE_dict}.   A number of image classification and restoration algorithms have been proposed based on sparse representations.   In particular, sparse representation-based classification (SRC) algorithm \cite{wright2009robust} has gained a lot of traction.  The basic idea of SRC is to identify the correct class by seeking the sparsest representation of the test sample in terms of the training.  The SRC algorithm was originally proposed for face recognition and later extended for iris recognition and automatic target recognition in \cite{iris_SRC} and \cite{SRC_ATR}, respectively.  A simultaneous dimension reduction and classification framework based on SRC was proposed in \cite{SRC_dim_reduction}.    Furthermore, non-linear kernel extensions of the SRC method have also been proposed in \cite{Kernel_SRC}, \cite{Hien_TIP_KKSVD}, \cite{SR_kenrel}, \cite{SRC_MKL}.

The SRC algorithm and its variants are essentially based on the \emph{closed world assumption}.  In other words, it is assumed that the testing data pertains to one of $K$ classes that are used during training.  But in practice, testing data may come from a class that is not necessarily seen in training.  This problem where the testing data corresponds to a class that is not seen during training is known as \emph{open set recognition} \cite{scheirer2013toward}.  Consider the problem of animal classification.  If the training samples correspond to $K$ different animals, then given a test image corresponding to an animal from one of the $K$ classes, the algorithm should be able to determine its identity.  However, if the test image corresponds to an animal which does not match one of the $K$ animals seen during training, then the algorithm should have the capability to ignore or reject the test sample \cite{animal_openset}.  

The goal of an open set recognition algorithm is to learn a predictive  model that classifies the known data into correct class and rejects the data from open class.  As a result, one can view open set recognition as tackling both the classification and novelty detection problem at the same time.  Novelty detection refers to the problem of finding anomalous behaviors that are inconsistent with the expected pattern.  A novelty detection problem can be formulated as a hypothesis testing problem where the null hypothesis, $\mathcal{H}_0$, implies the test sample coming from normal class and the alternative hypothesis, $\mathcal{H}_1$, indicates the presence of anomalies and the objective is to find the best  threshold that separates $\mathcal{H}_0$ from $\mathcal{H}_1$.  

A number of approaches have been proposed in the literature for open set recognition.  For instance, \cite{scheirer2013toward}  introduced a concept of \emph{open space risk} and developed a 1-vs-Set Machine formulation using linear SVMs for open set recognition.  In \cite{Scheirer_2014_TPAMIb}, the concept of Compact Abating Probability (CAP) was introduced for open set recognition.  In particular, Weibull-calibrated SVM (W-SVM) algorithm was developed which essentially combines the statistical Extreme Value Theory (EVT) with binary SVMs for open set recognition.  Also, the W-SVM framework was recently used in \cite{openset_spoof} for fingerprint spoof detection.  In \cite{camera_openset}, an open set recognition-based method was developed to identify whether  or  not  an  image  was  captured  by  a  specific digital  camera.

\begin{figure*}[htp!]
\centering
\includegraphics[width=1\textwidth]{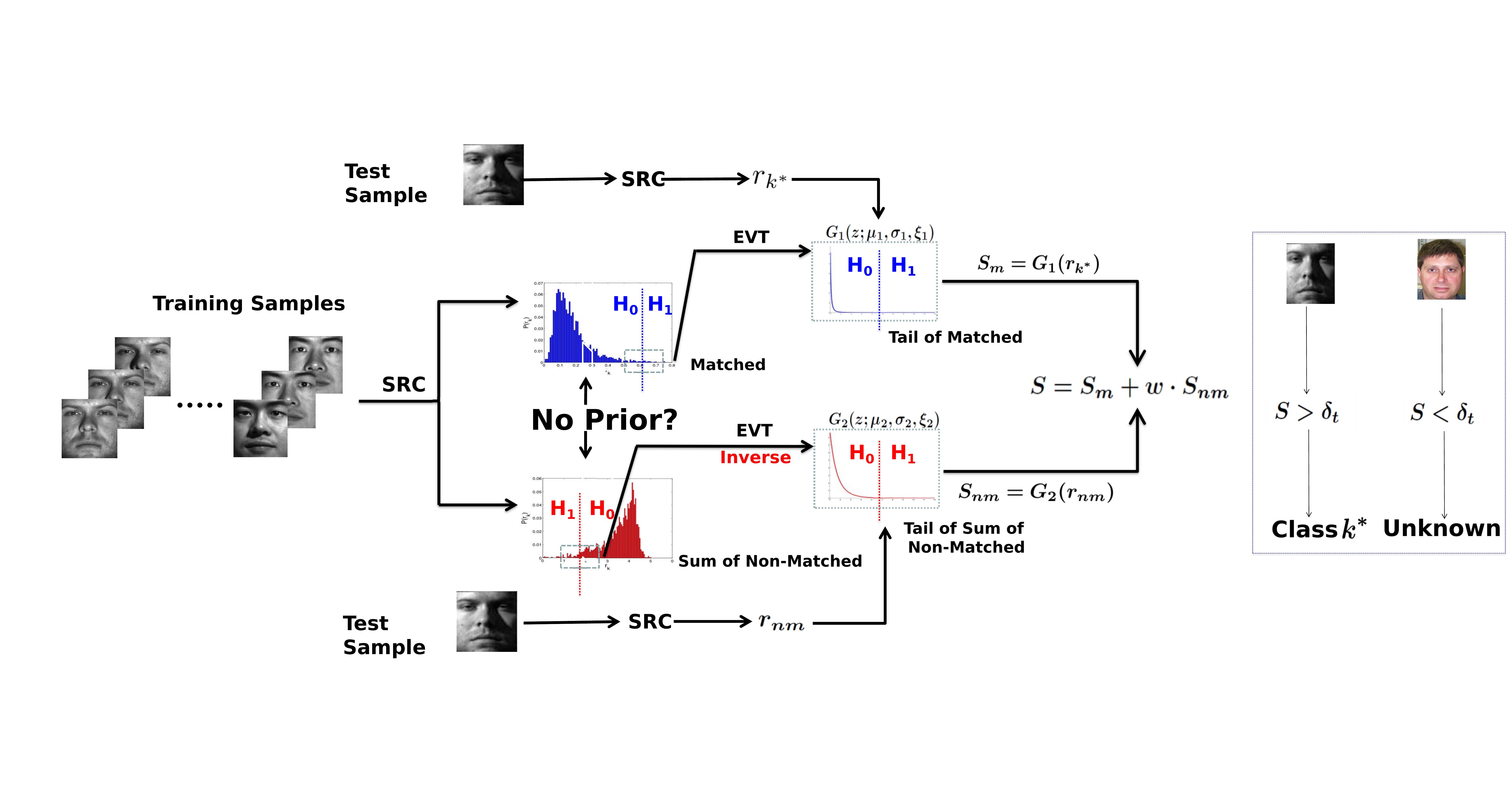} 
\vskip-30pt
\caption{Overview of the proposed SROSR algorithm.  Given training samples, we model tail part of the matched reconstruction error distribution and the sum of non-matched reconstruction error using the statistical EVT. Given a novel test sample, the modeled distributions and the matched and the sum of non-matched
reconstruction errors are used to calculate the confidence scores. Then, these scores are fused to obtain the final score for recognition.}\label{fig:overview}
\end{figure*}

In order to reject invalid samples, the notion of Sparsity Concentration Index (SCI) was proposed in \cite{wright2009robust}.  Similarly, a rejection rule based on the ratio of the first two highest projection scores was developed for rejecting non-face images in \cite{patel2012dictionary}.  The rejection rules defined using sparse representations in \cite{wright2009robust} and \cite{patel2012dictionary} were specifically designed to reject non-face images.  As will be shown later, these rules do not work well on general open set recognition problems.

In this paper, we extend the SRC formulation for open set recognition.  Our method relies on the statistical EVT \cite{pickands1975statistical} and consists of two main stages.  In the first stage, the tail distributions of the matched reconstruction errors and the sum of non-matched reconstruction errors are modeled using the EVT to simplify the open set recognition problem into two hypothesis testing problems.  In the second stage, the reconstruction errors corresponding to a test sample from each class are calculated and the confidence scores based on the two tail distributions are fused to determine the identity of the test sample.  Figure~\ref{fig:overview} gives an overview of the proposed Sparse Representation-based Open Set Recognition (SROSR) algorithm.

This paper is organized as follows.  In Section~\ref{sec:background}, we give a brief background on the EVT and the SRC algorithm.  Details of the proposed SROSR algorithm are given in  Section~\ref{sec:SROSR}.   Experimental results are presented in Section~\ref{sec:results} and Section~\ref{sec:conclusion} concludes the paper with a brief summary and discussion.

\section{Background}\label{sec:background}
In this section, we review some related work in SRC and EVT.  

\subsection{Sparse Representation-based Classification}
Stack the training samples from the $i$-th class as columns of a large matrix $\mathbf{Y}_i\in \R^{M\times N_{i}}$, and use $$ \mathbf{Y} = [\mathbf{Y}_1, \mathbf{Y}_2, \dots, \mathbf{Y}_K]\in \R^{M\times N},$$ as the dictionary of training samples from $K$ classes, where $N=\sum_{i}N_{i}$ is the total number of training samples and $M$ is the dimension of each training sample.  Let $\mathcal{L}^{Y}$ denote the corresponding label set.   If the $\mathbf{Y}_i$ are sufficiently expressive \cite{zhang2013toward}, a new input sample from the $i$-th class, stacked as a vector $\mathbf{y}_t \in \R^M$, will have a sparse representation $$\mathbf{y}_t = \mathbf{Y} \mathbf{x}$$ in terms of the training data $\mathbf{Y}$: $\mathbf{x}$ will be nonzero only for those samples from class $i$.   The sparse coefficient vector $\mathbf{x}\in \R^{N}$ can be estimated by solving the following optimization problem
\begin{equation}\label{eq:BP}
\hat{\mathbf{x}}=\arg\min_{\mathbf{x}} \|\mathbf{x}\|_{1}\;\;\text{s.t.}\;\;  \|\mathbf{y}_{t}-\mathbf{Y}\mathbf{x}\|_2<\epsilon,
\end{equation}
where we have assumed that the observations are noisy with noise energy $\epsilon$ and $\|\mathbf{x}\|_{1}=\sum_{i}|x_{i}|$.   The sparse code $\hat{\mathbf{x}}$ can then be used to determine the class of $\mathbf{y}_t$ based on the class residuals
\begin{equation}\label{eq:res}
r_k = \| \mathbf{y}_t - \mathbf{Y}_k \hat{\mathbf{x}}_k\|_2,\;\;\;k=1,\ldots , K,
\end{equation}
 where $\hat{\mathbf{x}}_{k}$ is the part of $\hat{\mathbf{x}}$ that corresponds to class $k$. Finally, the class $k^{*}$ that is associated to the test sample $\mathbf{y}_{t}$, can be declared as the one that produces the smallest approximation error
$$k^{*}=\text{class of }\mathbf{y}_{t}=\arg\min_{k}r_k.$$  This method provides excellent performance on several image classification datasets \cite{wright2009robust}, \cite{iris_SRC}, and is provably robust to errors and occlusion \cite{Wright2008-IT}.  The basic SRC algorithm is summarized in Algorithm~\ref{alg:SRC}.

 \begin{algorithm}
    \caption{Sparse Representation-based Classification}
    \label{alg:SRC}
    \begin{algorithmic} 
    \REQUIRE $\mathbf{Y}$, $\mathcal{L}^{Y}$, $\epsilon$, $\mathbf{y}_{t}$   
   \\ 
\STATE $\hat{\mathbf{x}} = \arg\min_{\mathbf{x}} \|\mathbf{x}\|_{1}\;\;\text{s.t.}\;\;  \|\mathbf{y}_{t}-\mathbf{Y}\mathbf{x}\|_2<\epsilon$
\STATE $r_k = \| \mathbf{y}_t - \mathbf{Y}_k \hat{\mathbf{x}}_k\|_2$ for $k=1, \ldots , K$
\STATE $k^{*}=\arg\min_{k}r_k$

              \ENSURE $k^{*},\mathbf{r}=[r_{1}, r_{2}, \ldots , r_{K}]$
  \end{algorithmic}
  \end{algorithm}

In order to reject outliers, the following SCI rule was defined in \cite{wright2009robust}
\begin{equation}\label{eq:SCI}
\text{SCI}(\mathbf{x})=\frac{\frac{K \times \max_k{\|\mathbf{x}_{k}\|_{1}}}{\|\mathbf{x}\|_{1}}-1}{K-1}\in [0,1].
\end{equation}  Sparsity coefficient index takes values between 0 and 1.  The SCI values close to 1 correspond to the case where the test image can be approximately represented by using only images from a single class.  If the SCI value of the recovered coefficient is close to zero, then the coefficients are spread across  all  classes. Hence,  the  test  vector  is  not  similar  to  any  of the classes and can be rejected.  A threshold can be chosen to reject invalid test samples if $\text{SCI}(\hat{\mathbf{x}})<\alpha$ and otherwise accepted as valid, where $\alpha$ is some chosen threshold between 0 and 1.

 \subsection{Extreme Value Theory}
Extreme value theory is a branch of  statistics analyzing the distribution of data of abnormally high or low values. It  has been applied in Finance \cite{kotz2000extreme}, Hydrology \cite{smith1989extreme} and novelty detection problems \cite{roberts1999novelty}, \cite{clifton2011novelty},  \cite{EVT_iccv_railway}.  In this section, we give a brief overview of the statistical EVT.
 
Assume that we are given $n$ i.i.d. samples $\{Z_1,Z_2,...,Z_n\} $\quad drawn from an unknown  distribution $F(z)$.  Denote $$Z_m=\max_i {Z_i} \quad i\in[1,n].$$  The Fisher-Tippett-Gnedenko theorem \cite{fisher1928limiting} states that if there exists a pair of parameters $(a_n,b_n)$, subject to the condition  $a_n>0$ and $b_n \in \R$, then 
 \begin{equation}
 \lim\limits_{n \to \infty} P\left(\frac{Z_m-b_n}{a_n}\right)=E(z),
 \end{equation}
 where $E(z)$ is a non-degenerate distribution that belongs to either Fr{\'e}chet, Weibull or Gumbel distribution.  These distributions can be represented as a Generalized Extreme Value distribution (GEV) as follows
 \begin{equation}
E(z;\mu,\sigma,\xi)=\exp^{-p(z)},
 \end{equation}
 where $$p(z)=\left(1+ \xi\left(\frac{z-\mu}{\sigma}\right)\right)^{-1/\xi}$$
and  $\mu, \sigma$ and $\xi$ are the location, scaling and shape parameters, respectively.

There are two challenges that one has to overcome before using the GEV distribution to model the tail distribution of data.  Firstly, we have to choose which distribution to use among the three based on prior knowledge. Secondly, we need to segment the data into several parts and model the maximum in each part as a distribution using GEV. However,  to overcome these challenges, an alternative method based on the Generalized Pareto distribution (GPD), denoted as $G(z)$\footnote{Here, $G$ is the Cumulative Distribution Function (CDF) of the GPD.}, was proposed in \cite{pickands1975statistical} to estimate the tail distribution of data samples. It was shown that given a sufficiently large threshold $u$, 
the probability of an observation exceeding $u$ by $z$ conditioned on $u$ can be approximated by 
 \begin{equation}
\begin{split}
\lim_{n \rightarrow \infty} P(Z>z+u|Z>u)=1-G(z),
\end{split}
 \end{equation}
 with $$G(z)=1-\left(1+\xi \frac{z}{\sigma}\right)_{+}^\frac{-1} {\xi}, \;\;z>0,$$ 
 where $\sigma >0$, $\xi \in \R$ and $x_{+}=\max(x,0)$. 
 
To estimate the parameters of GPD, one can use the maximum likelihood estimation (MLE) method introduced in \cite{gpdconvergence1}. Even though there is the possibility that the parameters of GPD don't exist and that maximum likelihood estimation may not converge when $\xi >1/2$, it has been shown that these are extremely rare cases in practice \cite{gpdconvergence1} \cite{gpdgoodness} . 

\section{Sparse Representation-based Open-Set Recognition (SROSR)}\label{sec:SROSR}
In \cite{scheirer2013toward} the notion of ``Open set Risk" was defined as the cost of labeling the open set sample as known sample.  Based on this, one can minimize the following cost to develop an open set recognition algorithm 
\begin{equation}\label{eq:openset_risk}
\arg\min_{f} {C_o(f)+\lambda _r C_\epsilon(f)},
\end{equation}
where $f$ is a measurable function, $C_o(f)$ denotes open set risk, $C_\epsilon(f)$ denotes empirical risk for classification and  $\lambda _r$ is a parameter  that balances open set risk and empirical risk.

\begin{figure}[htp!]
\begin{center}
\includegraphics[width=9.5cm, height=6.2cm]{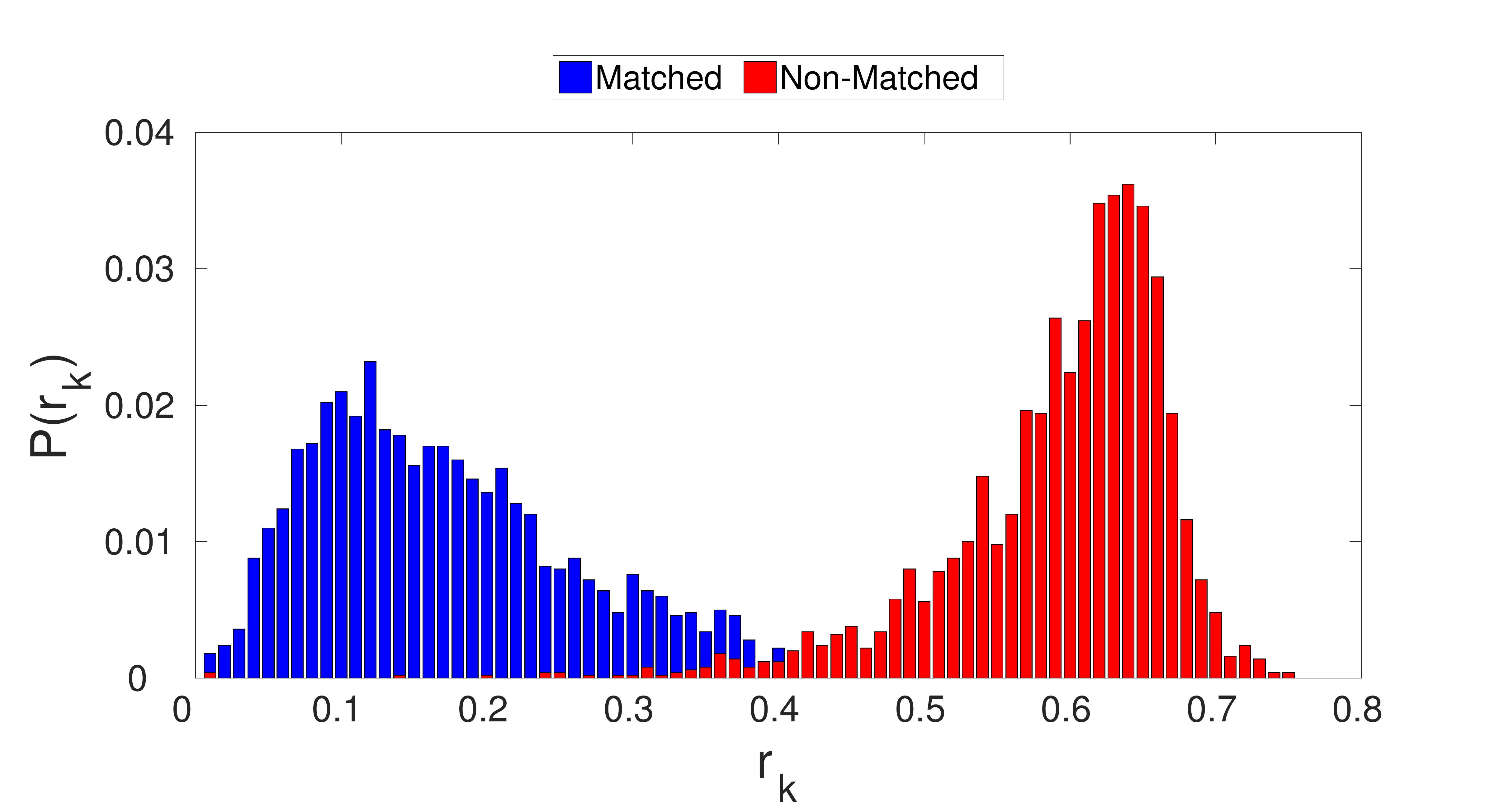} 
\caption{Histogram of the matched and non-matched reconstruction errors.  Matched reconstruction errors are the errors corresponding to the sparse coefficients of digit 9 and non-matched reconstruction errors are the errors that are generated by the sparse coefficients of all other digits when training samples consists of digits 0 to 9 and the test samples correspond to digit 9. All samples are from the MNIST dataset.}\label{fig:matched_vs_nonmatched}
\end{center}
\end{figure}

The SRC algorithm uses residuals from \eqref{eq:res} for classification which can also be used to model $f$ in  \eqref{eq:openset_risk} for open set recognition.  This is due to the following reason.  If the test sample corresponds to class $k$, then the reconstruction error corresponding to class $k$ should be much lower than that corresponding to the other classes.  As a result, there may be a distinction between matched and non-matched reconstruction errors. To illustrate this, we plot the distributions of matched and  non-matched  reconstruction errors using the samples from the MNIST handwritten digits dataset \cite{Mnist}, shown in Figure~\ref{fig:matched_vs_nonmatched}.  Training samples consists of digits 0 to 9 and test samples correspond to digit 9.   Matched reconstruction errors here mean that the errors correspond to the sparse coefficients of digit 9 and non-matched reconstruction errors mean that the errors are generated by the sparse coefficients of all other digits.  One can see from this figure that matched classes' reconstruction errors  follow some underlying distribution.  If one can fit a probability model $P(r_k)$ to describe the distribution of the reconstruction errors of the matched class, then one can reformulate the open-set recognition problem as a hypothesis testing for novelty detection problem as
\begin{align}\label{eq:hypo}
\nonumber \mathcal{H}_0&: P(r_k) \le \delta \\
\mathcal{H}_1&: P(r_k)>\delta,
\end{align}
where the null hypothesis $\mathcal{H}_0$ implies that the test data are generated from the distribution $P(r_k)$, and the alternative hypothesis $\mathcal{H}_1$ implies that test data correspond to the classes other than the ones considered in training and $\delta \in [0,1]$ is the threshold for rejection.

However, as we have no prior knowledge on the underlying distribution of the matched reconstruction errors, we cannot fit a proper distribution on them.  Instead, we can apply the EVT on the tail of the matched distribution as we are only concerned about the right tail of this distribution for hypothesis testing.  As the implementation of GEV on real data is difficult, we instead use the GPD to model the tail of the matched distribution.
Once we learn the distribution of the tail, we can modify the  hypothesis testing problem Eq. \ref{eq:hypo} to the following 
 \begin{align}
\nonumber \mathcal{H}_0: G(r_k)& \le \delta_g \\
\mathcal{H}_1: G(r_k)&>\delta_g,
 \end{align}
where $G(r_k)$ is the learned GPD distribution for fitting the right tail of $r_k$ and $\delta_{g}$ is the rejection threshold.

\begin{figure}[t]
\begin{center}
\includegraphics[width=9.5cm, height=6.5cm]{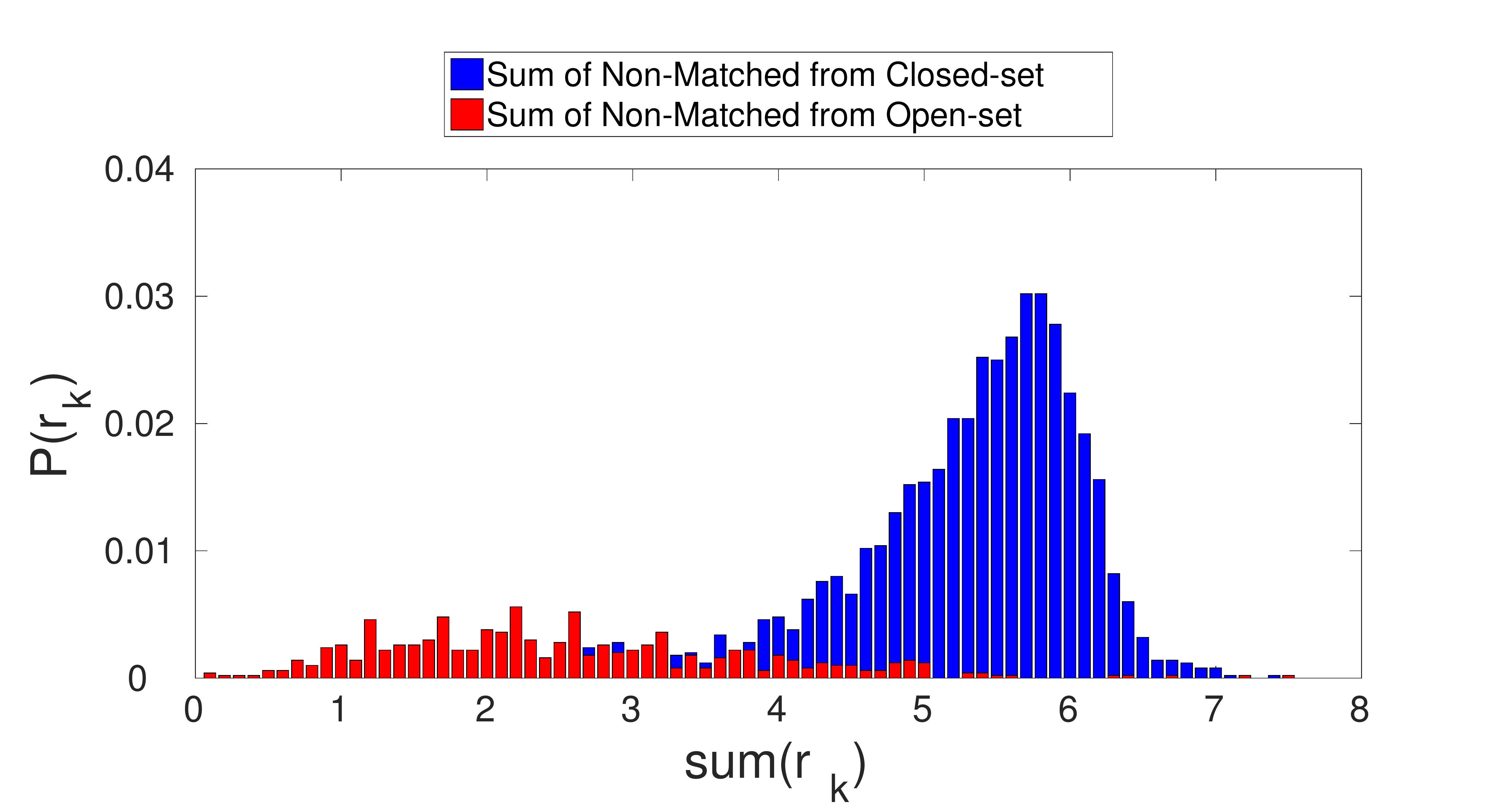} 
\caption{Histogram of the sum of non-matched reconstruction errors corresponding to the closed set classes 0 to 5 and the sum of non-matched reconstruction errors corresponding to the open set digits 6 to 9 . All samples are from MNIST dataset.}\label{fig:sum_matched}
\end{center}
\end{figure}

When SRC is used for classification, we don't only get the information of the matched reconstruction errors but we also have access to the non-matched reconstruction errors which can be used to enhance the performance of our open set recognition algorithm.   Due to the self expressiveness property of the SRC algorithm \cite{wright2009robust}, the sparse coefficients corresponding to open set samples are very different from that of the closed set samples and they follow a certain pattern.   If an open set sample is written as a linear combination of the training samples from closed set then the resulting sparse coefficient vector will not concentrate on any class but instead spread widely across the entire closed training set.  Thus, the distribution of the estimated sparse coefficient contains important information about the validity of open set sample.   In order to illustrate this point, we conduct the following toy experiment using the digits from the MNIST dataset.    Suppose that the training data only contains digits 0 to 5 and the test samples consist of closed set digits 0 to 5 and open set digits 6 to 9.   In  Figure~\ref{fig:sum_matched}, we plot the sum of the non-matched reconstruction errors corresponding to the closed set digits 0 to 5 and the sum of non-matched reconstruction errors corresponding to the open set digits 6 to 9.  As one can see from this figure that the sum of the non-matched reconstruction errors from the closed set digits 0 to 5 also follow a certain distribution that is very different from the distribution that one obtains from the errors corresponding to the open set digits.

As a result, we can formulate another hypothesis testing problem similar to \eqref{eq:hypo} for the sum of non-matched reconstruction errors. We can combine the two hypothesis testing problems together to make the open set recognition algorithm more accurate.  As we are only interested in  the right tail of the matched distribution and the left tail of the sum of non-matched distribution, we apply an inverse procedure to the random variable $Z$ as $$Z_I=-Z.$$  So the right tail of $Z_I$ is the left tail of $Z$.

\subsection{Training}
In the training phase, we have to estimate the parameters for fitting the tail distribution based on the GPD.  Estimating the parameters based on MLE requires the availability of multiple reconstruction errors.  To deal with this issue, we propose the following iterative procedure.    In each iteration, we first randomly order the training samples from each class $\mathbf{Y}_{i}$ and then partition them into two sets - cross-train $\mathbf{Y}_{i}^{tr}$  and  cross-test $\mathbf{Y}_{i}^{te}$.  Samples in the cross-train set $\mathbf{Y}_{i}^{tr}$ and samples in the cross-test set $\mathbf{Y}_{i}^{te}$ are used as training and testing samples, respectively for the SRC algorithm during this particular iteration.   The cross-test and cross-train sets contain 20 and 80 percent of the training samples in $\mathbf{Y}_{i}$, respectively.  Let $\mathcal{L}^{tr}_{i}$ and $\mathcal{L}^{te}_{i}$ denote the associated label sets corresponding to $\mathbf{Y}_{i}^{tr}$ and $\mathbf{Y}_{i}^{te}$, respectively.  Once the training samples from all classes are partitioned into cross-train and cross-test sets, combine the cross-train samples from all $K$ classes into a cross-train matrix $\mathbf{Y}^{tr}=[\mathbf{Y}^{tr}_{1}, \mathbf{Y}^{tr}_{2}, \ldots , \mathbf{Y}^{tr}_{K}]$ and their associated labels into a label set $\mathcal{L}^{tr}=\{\mathcal{L}^{tr}_{1}, \mathcal{L}^{tr}_{2}, \ldots , \mathcal{L}^{tr}_{K}\}$.  Similarly, combine the cross-test sets into a cross-test matrix $\mathbf{Y}^{te}=[\mathbf{Y}^{te}_{1}, \mathbf{Y}^{te}_{2}, \ldots , \mathbf{Y}^{te}_{K}]$ and their labels into a label set $\mathcal{L}^{te}=\{\mathcal{L}^{te}_{1}, \mathcal{L}^{te}_{2}, \ldots , \mathcal{L}^{te}_{K}\}$.   Use $(\mathbf{Y}^{tr}, \mathbf{Y}^{te}, \mathcal{L}^{tr}, \mathcal{L}^{te}, \epsilon)$ as the inputs to the SRC algorithm and obtain the reconstruction error vector $\mathbf{r}_i$. We repeat this process for $L$ times and gather the matched $\mathbf{R}_{i}^{m}$ and the sum of non-matched reconstruction errors  $\mathbf{R}_{i}^{nm}$, respectively for $i=1, \ldots , K$, for fitting the tail distribution based on the GDP.   The entire training phase of our method is summarized in Algorithm~\ref{alg:training}, where $\rho$ indicates the tail size.

\begin{algorithm}
    \caption{Pseudocode for SROSR Training}
    \label{alg:training}
    \begin{algorithmic} 
    \REQUIRE $\mathbf{Y}, \rho, \epsilon,L$, $\mathcal{L}^{Y}$   
   \\ Initialization
\FOR{$i=1:K$}  
\FOR{$j=1:L$}  
\STATE $\tilde{\mathbf{Y}}_{i}$ = randomly ordered $\mathbf{Y}_{i}\in \R^{M\times N_{i}}$
 \STATE $ N_{tr}=N_i \times 0.8$
 \STATE 
 $\mathbf{Y}_{i}^{tr} = \tilde{\mathbf{Y}}_{i}(:,1:N_{tr})$
 \STATE
$\mathcal{L}^{tr}_{i} = \text{Labels of }\mathbf{Y}_{i}^{tr}$
 \STATE 
 $\mathbf{Y}_{i}^{te} = \tilde{\mathbf{Y}}_{i}(:, N_{tr}+1:\text{end})$
  \STATE
 $ \mathcal{L}^{te}_{i} = \text{Labels of }\mathbf{Y}_{i}^{te}$
 \\$\mathbf{r}_{i}(j,:)$ $\leftarrow$ \text{SRC} $(\mathbf{Y}^{tr}, \mathbf{Y}^{te}, \mathcal{L}^{tr}, \mathcal{L}^{te}, \epsilon)$
  \\
    \ENDFOR  
    
 $\mathbf{R}_i^m$ = $[\mathbf{r}_{i}(1,i), \ldots  ,\mathbf{r}_{i}(L,i)]$
    \\
 $\mathbf{R}_i^{nm}$ = $[\sum_{p:p\ne i} \mathbf{r}_{i}(1,p), \ldots   , \sum_{p:p\ne i} \mathbf{r}_{i}(L,p)]$
    \\
$\boldsymbol{\sigma}_m(i), \boldsymbol{\xi}_m(i)$ $\leftarrow$ GPDfit($\mathbf{R}^m,  \rho$)
\\
  $\boldsymbol{\sigma}_{nm}(i), \boldsymbol{\xi}_{nm}(i)$ $\leftarrow$ GPDfit($ \mathbf{-R}^{nm}, \rho$)
     \ENDFOR 
              \ENSURE $\boldsymbol{\sigma}_m, \boldsymbol{\xi}_m,  \boldsymbol{\sigma}_{nm}, \boldsymbol{\xi}_{nm}$
  \end{algorithmic}
  \end{algorithm}

\subsection{Testing}
Given a novel test sample $\mathbf{y}_{t}$, we compute its sparse coefficient $\hat{\mathbf{x}}$ by solving the $\ell_{1}$-minimization problem Eq. \ref{eq:BP}.  We then obtain $K$ reconstruction errors as required by the SRC algorithm.  We choose the class with the minimum reconstruction error as the candidate class.  We then obtain two probability scores by fitting matched and sum of non-matched reconstruction errors to their corresponding GPDs.   As the two raw reconstruction errors are all normalized into probabilities by their corresponding GPDs, we can add the two probability scores together with appropriate weights to obtain the final score.  We set the weight, $w$, as  
$$w=\frac{1}{3}(1-\text{Openness}),$$
where \begin{equation}
\text{Openness}=1-\sqrt{ \frac {2\times N_{TA}}{N_{TG}+N_{TE} }},
\end{equation}
and $N_{TA},  N_{TG}$ and $N_{TE}$ are  the number of training classes, the number of target classes to be identified, and the number of testing classes, respectively \cite{scheirer2013toward}.   If  `Openness = 0', then our setting reduce to the traditional classification problem (i. e., a completely closed problem). With the growth of `Openness', more and more unknown classes will appear during testing.  As a result, the weight on the non-matched probability scores will decrease.   

Our testing algorithm is summarized in Algorithm~\ref{alg:testing}.  The inputs required during testing are the test sample $\mathbf{y}_t$, training samples $\mathbf{Y}$, the estimated parameters for the matched $(\boldsymbol{\sigma} _m, \boldsymbol{\xi}_m)$ and the sum of non-matched distributions $(\boldsymbol{\sigma}_{nm}, \boldsymbol{\xi} _{nm})$, rejection threshold $\delta _t$ and the weight $w$. The output of the testing phase is one of the following classes $\{1, 2, \ldots , K, \mathcal{O}\}$, where $\mathcal{O}$ represents the open class. 

 \begin{algorithm}
    \caption{Pseudocode for SROSR Testing}
    \label{alg:testing}
    \begin{algorithmic} 
    \REQUIRE $\mathbf{y}_{t}, \mathbf{Y}, \boldsymbol{\sigma}_m, \boldsymbol{\xi}_m,  \boldsymbol{\sigma}_{nm}, \boldsymbol{\xi}_{nm}, \delta _t, w,\epsilon$  
 
    1: $\mathbf{r}$ $\leftarrow$ \text{SRC} $(\mathbf{Y}, \mathbf{y}_{t}, \mathcal{L}^{Y},  \epsilon)$
    \\3: $k^{*} = \arg\min_{i}  {r}_{i}$ 
    \\4: $r_m={r}_{k^{*}}$, $r_{nm}=\sum_{i=1,i\ne k^{*}}^{K}{r}_{i} $
    \\5: ${S_{m}}= G(r_m; \boldsymbol{\sigma}_m (k^*), \boldsymbol{\xi}_m(k^*))$,
    \\ \quad ${S_{nm}}= G(r_{nm}; \boldsymbol{\sigma}_{nm} (k^*), \boldsymbol{\xi}_{nm}(k^*))$
    \\6: $S= S_m+w \ldots  S_{nm}$  
    \IF{$S>\delta _t$}  
    \STATE $\text{Class of } \mathbf{y}_{t}=\mathcal{O}$  
    \ELSE  
    \STATE $\text{Class of } \mathbf{y}_{t} = k^{*}$   
    \ENDIF 
     \ENSURE $k^{*}$ or $\mathcal{O}$
  \end{algorithmic}
  \end{algorithm}

\section{Experimental Results}\label{sec:results}
In this section, we present several experimental results demonstrating the effectiveness of the proposed SROSR method on open set recognition.  In particular, we present the open set recognition results on the MNIST handwritten digits dataset \cite{Mnist}, Extended Yale B face dataset \cite{Yaleb}, UIUC attribute dataset \cite{UIUC_attribute} and Caltech-256 dataset \cite{caltech_256}.  The comparison with other existing open set recognition methods such as 1-vs-All Multi-class RBF SVM with Platt Probability Estimation \cite{svmpro} and Pairwise Multi-class RBF SVM \cite{SVMcomparison}  in \cite{Scheirer_2014_TPAMIb} suggests that the W-SVM algorithm is among the best.  Hence,  we  treat  it  as  state-of-the-art  and  use  it  as  a  benchmark  for  comparisons  in  this  paper.    Furthermore, we compare the performance of our method with two other sparse representation based methods for rejecting invalid samples - SCI \cite{wright2009robust} and Ratio method  \cite{patel2012dictionary}.   Finally, we compare our method with a ``Naive" baseline where we estimate a reconstruction error threshold directly from training rather than using GPD to model the tail distributions.

Recognition accuracy and F-measure are used to measure the performance of different algorithms on open set recognition.  The F-meaure is defined as a harmonic mean of Precision and Recall
\begin{equation}
\text{F-measure}=2\cdot \frac{\text{Precision}\cdot \text{Recall}}{\text{Precision}+ \text{Recall}}, 
\end{equation}
where Recall is defined as 
$$\text{Recall}=\frac{\text{TP}}{\text{TP+FN}}$$
  and Precision defined as
   $$\text{Precision}=\frac{\text{TP}}{\text{TP+FP}}.$$ Here TP, FN, and FP denote true positive, false negative and false positive, respectively.  F-measure is always between 0 and 1.  The higher the F-measure the better the performance of an object recognition system.  Accuracy is defined as 
$$\text{Accuracy}=\frac{\text{TP}+\text{TN}}{\text{TN}+\text{TP}+\text{FP}+\text{FN}},$$ where TN denotes true negative.   The rejection threshold, $\delta_{t}$ was empirically determined.  In our experiments, we have used $\delta_{t}= 0.006\cdot(1+w), 0.007\cdot(1+w), 0.05\cdot(1+w),0.1\cdot(1+w) $ for the simulations with the MNIST dataset, Extended YaleB dataset, UIUC attribute dataset and  Caltech-256 dataset, respectively.   We choose the tail size $\rho$ based on cross-validation.  In particular, we set $\rho=0.14, 0.10, 0.39$ and $0.25$ for the experiments with the MNIST dataset, Extended YaleB dataset, UIUC attribute dataset and Caltech-256 dataset, respectively.  The noise level $\epsilon$ is set equal to $0.001$ for solving the SRC problem in our proposed SROSR framework.

 \subsection{Results on the Extended YaleB Dataset}
 The Extended Yale B Dataset consists of 2,414 frontal images of 38 individuals.  These images were
captured  under various controlled  indoor lighting  conditions.  Each class contains about 64 images.  They were cropped and normalized to  the  size  of  $32 \times 32$ pixels.  We randomly choose 10 classes for training and vary the openness by randomly selecting 10 to 28 classes.  The following steps summarize our data partition procedure on the Extended Yale B dataset.
\begin{enumerate}
\item Randomly select 10 classes among the 38 classes.
\item Randomly choose 80\% of the samples in each of the 10 selected classes as training samples.
\item  Select the remaining 20\% of the samples from step 2 and all the samples from the other 28 classes as testing samples.
\end{enumerate}
We repeat the above procedure 50 times and report the average F-measure and accuracy of different methods.

\begin{figure}[htp!]
\begin{center}
\includegraphics[width=4.2cm, height=4.2cm]{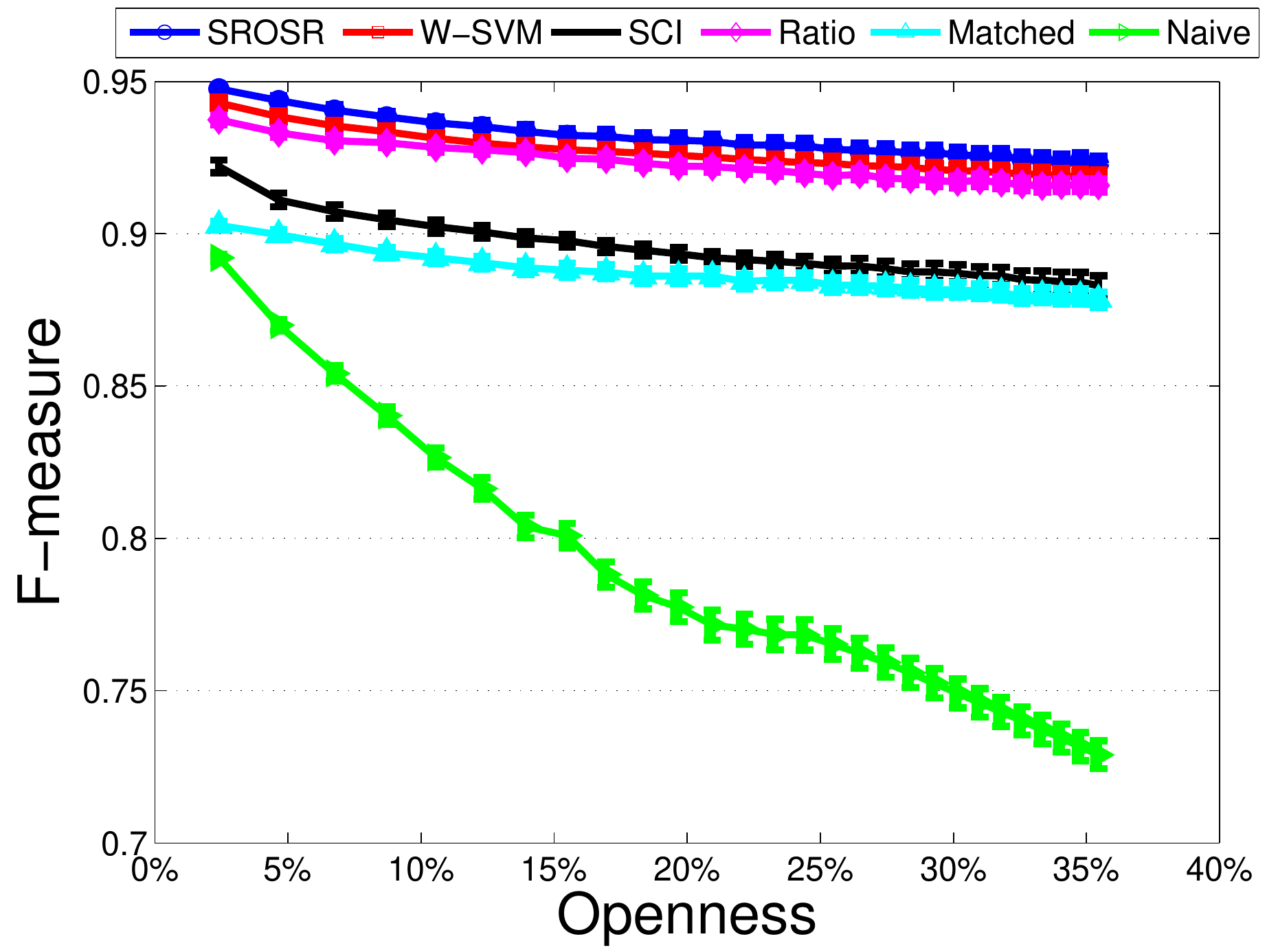}\includegraphics[width=4.2cm, height=4.2cm]{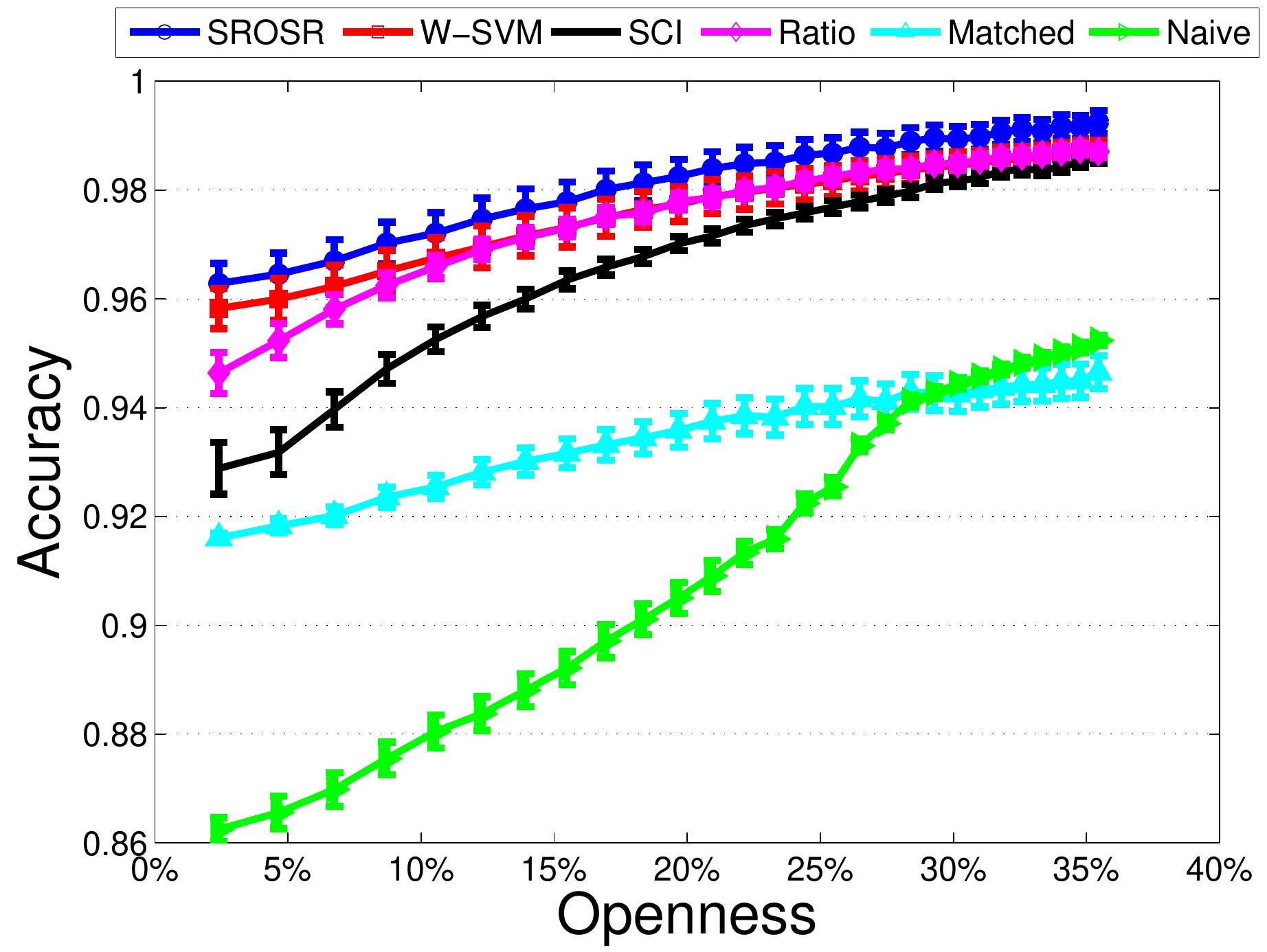}\\
(a)\hskip100pt(b)
\caption{Results on the Extended Yale B dataset.  (a) Openness vs F-Measure results.  (b) Openness vs Accuracy results.}\label{fig:yaleB_results}
\end{center}
\end{figure}

To show the significance of why we used the sum of non-matched reconstruction error distribution along with the matched error distribution, in this experiment, we consider just the matched reconstruction error distribution without fusing the sum of the non-matched reconstruction error distribution in our method.    The results as shown in Figure~\ref{fig:yaleB_results}.  Figure~\ref{fig:yaleB_results} (a) shows the average F-measure results on this dataset.  The face images in this dataset are cropped and well-aligned.  Furthermore, the images contain almost the same background.  As a result, all compared methods achieve very high F-measures on this dataset.   Figure~\ref{fig:yaleB_results} (b) shows the average accuracy of different methods as we vary openness.  As can be seen from both of these plots, the proposed SROSR method outperforms the other compared methods.  In particular, if only the matched reconstruction error distribution is considered, then the performance degrades significantly.  On the other had, when both the sum of non-matched and matched distributions are used, it greatly enhances the performance of the proposed SROSR algorithm.   This experiment clearly indicates that  both matched and the sum of non-matched reconstruction errors contain complementary information which can be used to improve the performance of an open set recognition algorithm.

 \subsection{Results on the MNIST Dataset}
 The MNIST dataset contains gray scale images of handwritten digits of size $28 \times 28$.  There are about 60,000 training images and 10,000 testing images corresponding to 10 classes in this dataset.  Following the experimental setting described in \cite{Scheirer_2014_TPAMIb}, we randomly choose 6 classes for training and alter the openness by the remaining 4 classes. We repeat this experiment 50 times and record the average F-measure and Accuracy.   Finally, we plot the Openness vs F-measure and Openness vs Accuracy curves to validate our approach.

\begin{figure}[htp!]
\begin{center}
\includegraphics[width=4.2cm, height=4.2cm]{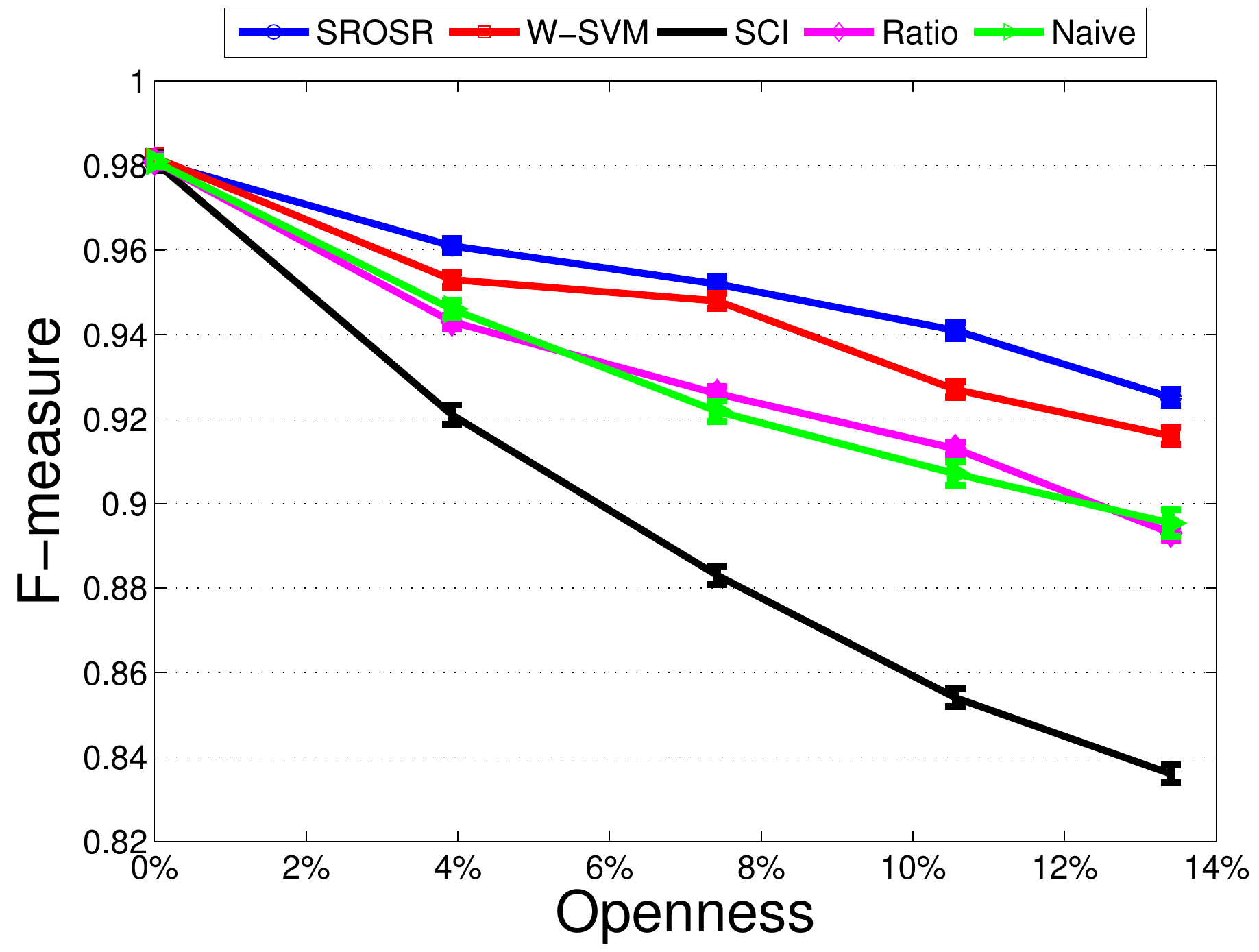}\includegraphics[width=4.2cm, height=4.2cm]{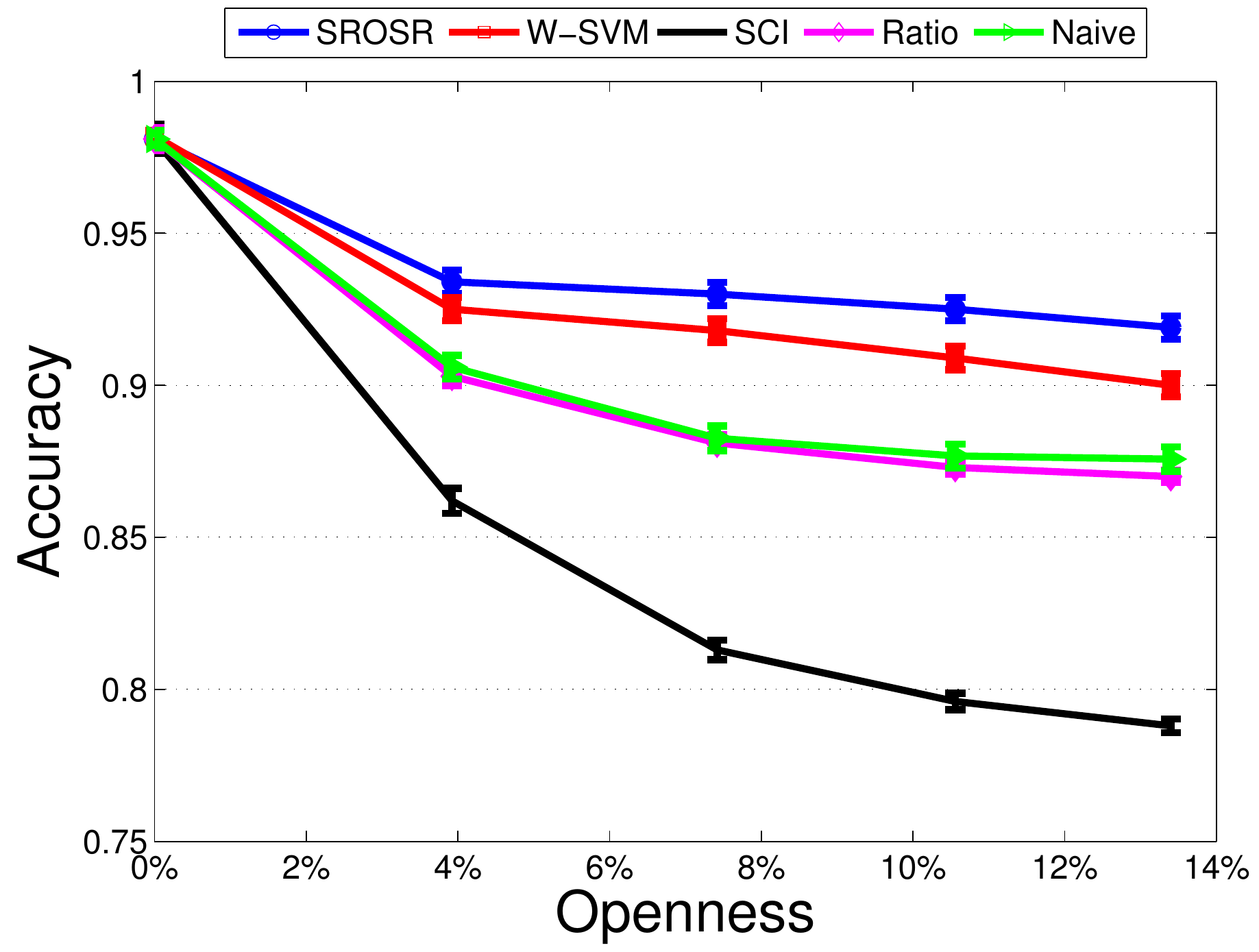}\\
(a)\hskip100pt(b)
\caption{Results on the MNIST dataset.  (a) Openness vs F-Measure results.  (b) Openness vs Accuracy results.}\label{fig:MNIST_results}
\end{center}
\end{figure}

The Openness vs F-measure and Openness vs Accuracy curves  corresponding to this experiment are shown in Figure~\ref{fig:MNIST_results} (a) and Figure~\ref{fig:MNIST_results} (a), respectively. It can be seen from these results that the proposed SROSR method performs better than the Naive method, the W-SVM method and the sparsity-based rejection methods.   Our method achieves the highest F-measure and accuracy among all the five methods as we vary openness.  The rejection methods such as SCI and Ratio are based on the sparsity of the test vector with respect to the training samples.  If an open set sample has a sparsity pattern similar to that corresponding to one of the training samples, then the SRC method based on SCI will not reject that sample.  This demonstrates that incorporating matched as well as non-matched reconstruction errors can significantly enhance the performance of a sparsity-based classification method on open set recognition.  

By comparing Figure~\ref{fig:yaleB_results} (b) with Figure~\ref{fig:MNIST_results} (b), we see that the accuracy in Figure~\ref{fig:yaleB_results} (b) increases while the accuracy in Figure~\ref{fig:MNIST_results} (b) decreases.  This is mainly due to the fact that the rejection accuracy is higher than the recognition accuracy on the Extended YaleB dataset while the rejection accuracy is lower than the recognition accuracy on the MNIST dataset.

 \subsection{Results on the UIUC Attribute Dataset}
\begin{figure}[htp!]
\begin{center}
\includegraphics[width=4.2cm, height=4.2cm]{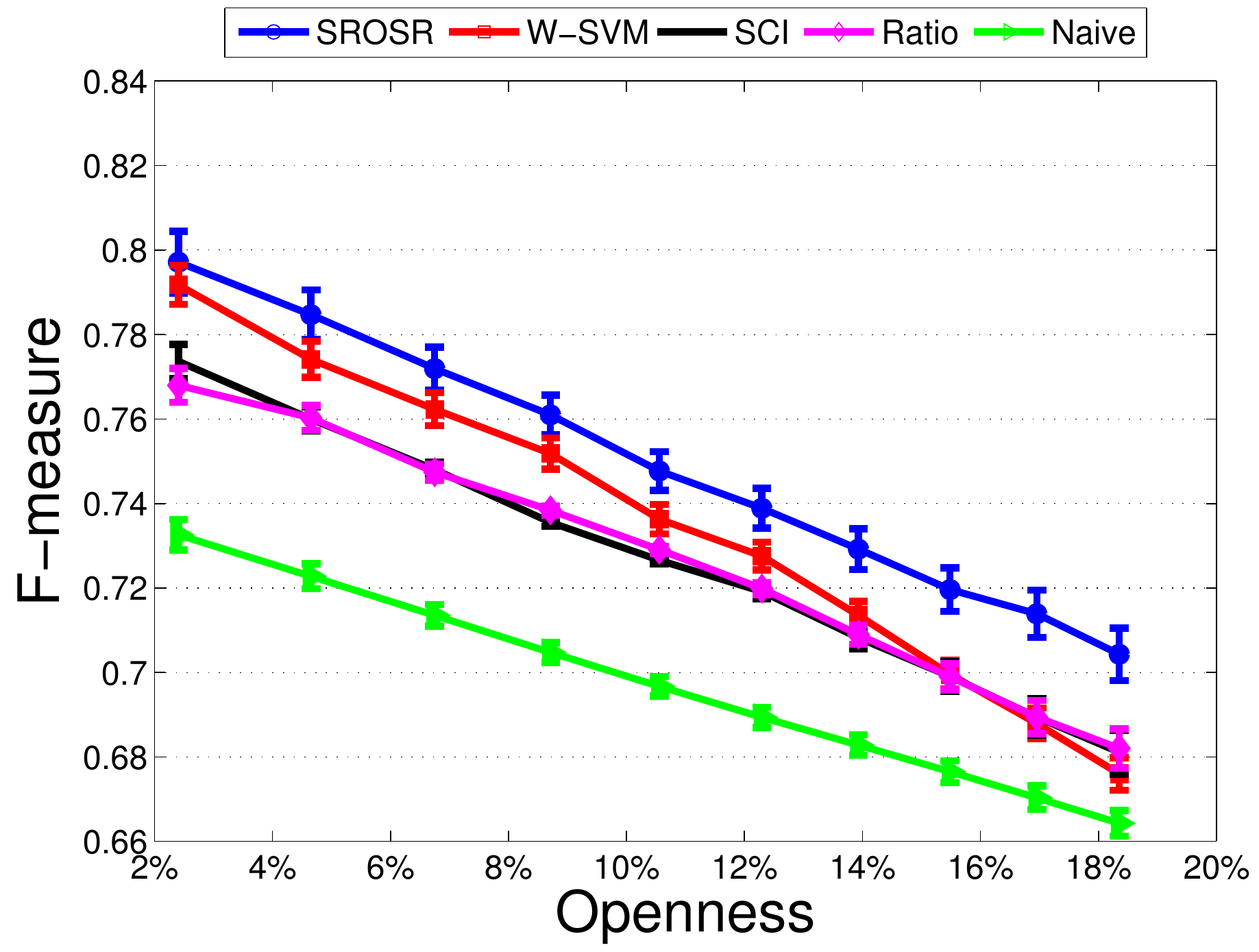}\includegraphics[width=4.2cm, height=4.2cm]{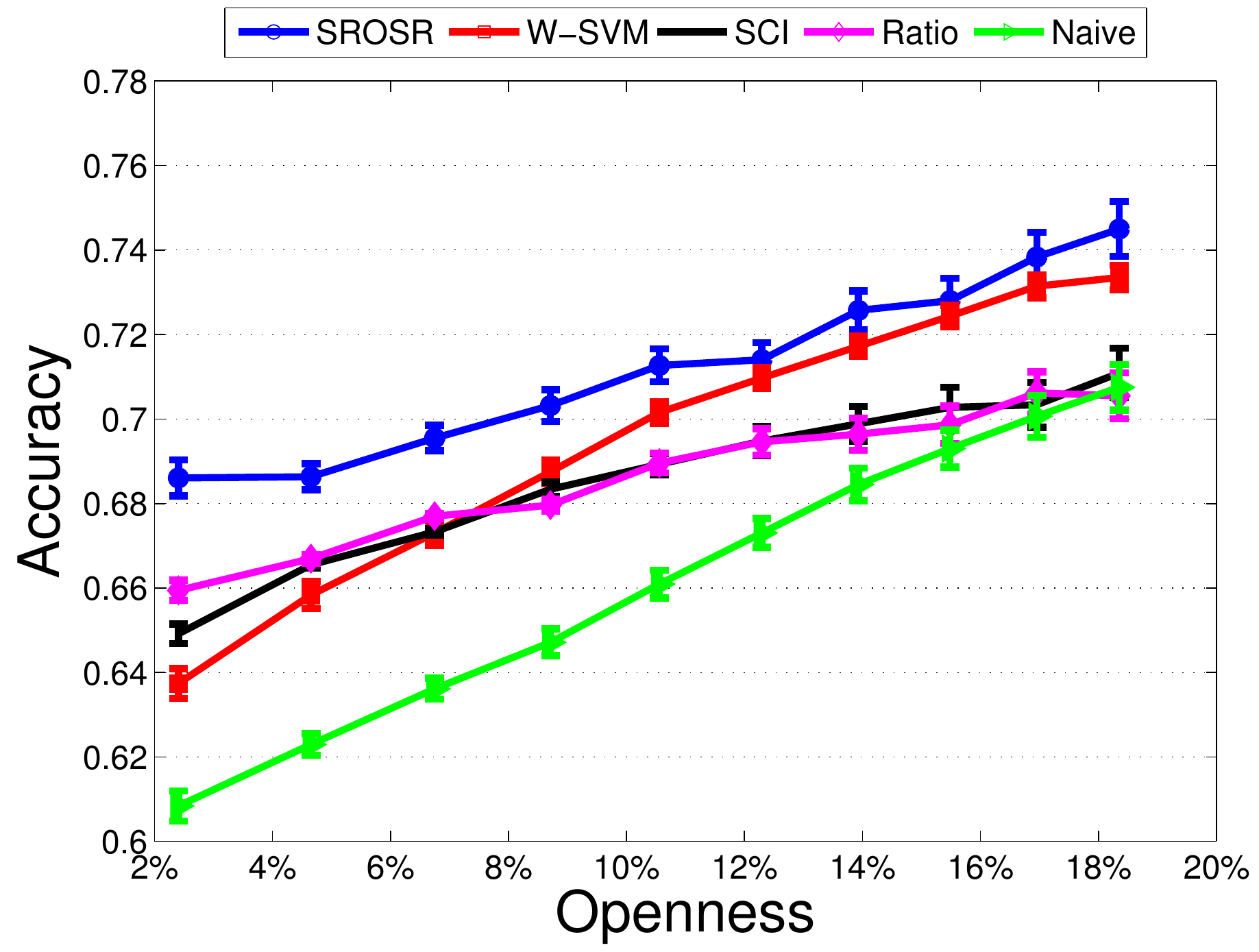}\\
(a)\hskip100pt(b)
\caption{Results on the UIUC attribute dataset.  (a) Openness vs F-Measure results.  (b) Openness vs Accuracy results.}\label{fig:UIUC_results}
\end{center}
\end{figure}

 The UIUC attributes dataset contains data in two parts - a-Pascal and a-Yahoo. The a-Pascal dataset has twenty object classes such as animals, vehicles, etc. and each category contains  150 to 1000 samples. The a-Yahoo dataset contains twelve additional object classes, which can be used as open set classes during testing.  We randomly choose 10 classes from the a-Pascal dataset as training and vary the openness by randomly selecting 1 to 10 classes from the a-Yahoo dataset. In each training class, we randomly choose 50 samples and in each testing class, we randomly choose 20 samples. We repeat the above procedure 50 times and average the F-measure and accuracy results.   Results are shown in Figure~\ref{fig:UIUC_results}.  As can be seen from this figure that SROSR outperforms the other methods.  In particular, as the openness is increased, our method can achieve much better F-measure and accuracies than the other compared methods.

\subsection{Results on the Caltech-256 Dataset}
\begin{figure}[htp!]
 \centering
\includegraphics[width=4.2cm, height=4.2cm]{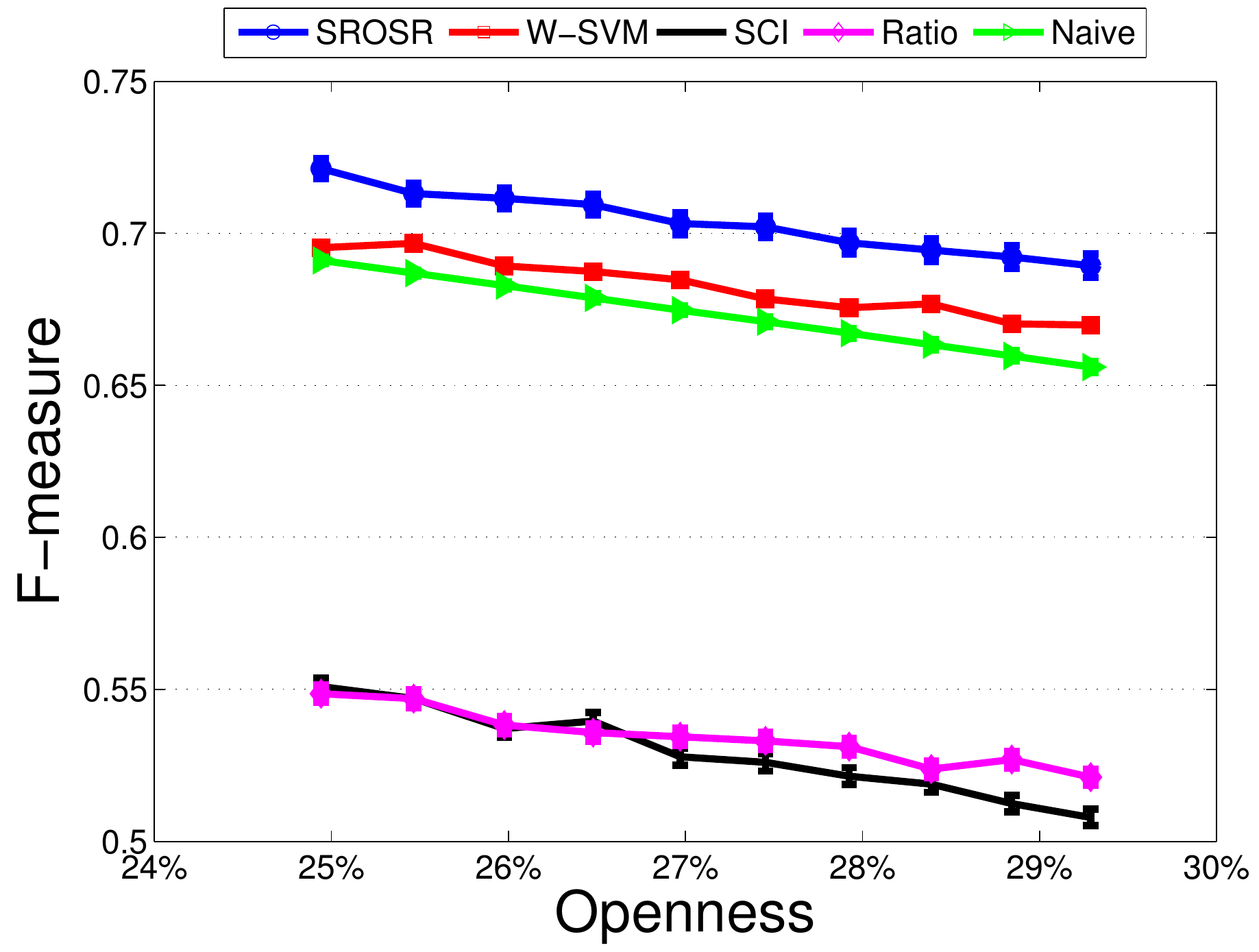}\hskip1pt\includegraphics[width=4.2cm, height=4.2cm]{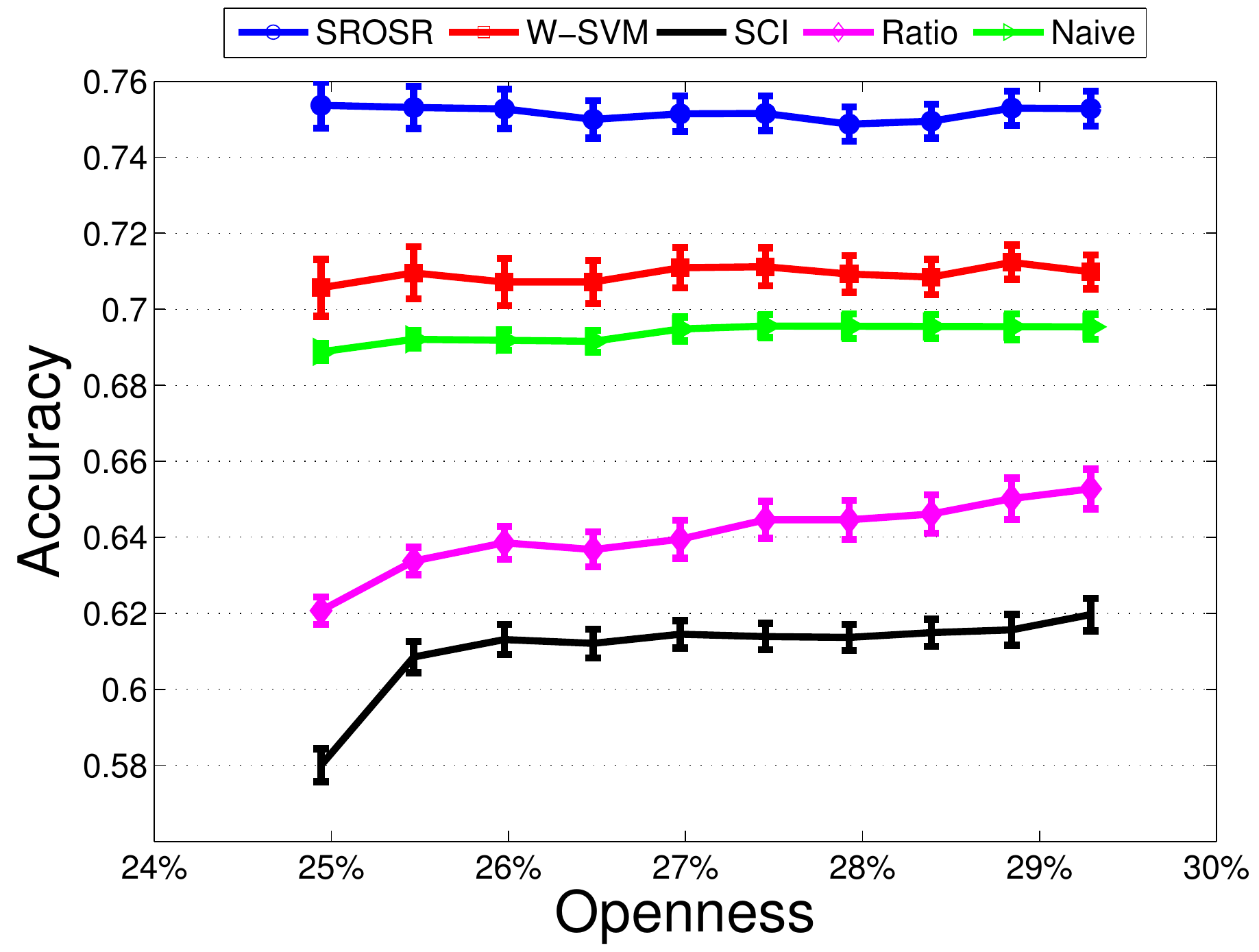}\\
(a)\hskip100pt(b)
\caption{Results on the Caltech 256 dataset.  (a) Openness vs F-Measure results. (b) Openness vs Accuracy results.}
\label{fig:results_caltech256}
\end{figure}

The Caltech-256 dataset contains 257 categories including one background \emph{clutter} class. Each category has about 80 to 827 images and most of the categories have about 100 images. In this experiment, we extracted the spatial pyramid features \cite{saptial_pyraid} from these images as input for all four methods.   The evaluation protocol  is very similar to the previous three experiments. We randomly select 20 categories as training classes and vary the openness by randomly selecting 31 to 40 classes out of the other 237 classes.  For all the selected classes, we randomly choose 50 samples for each training class and 20 samples for each testing class. So the openness of our experiments on the Caltech-256 dataset varies from 24.94\% to 29.29\%. We average the results over 50 random trails.  Figure~\ref{fig:results_caltech256} (a) and (b) show the average F-measure and accuracy curves of different methods as we vary the openness, respectively.  Overall, the proposed SROSR achieves the best  F-measure and accuracy results on this dataset compared to the other competitive open set recognition methods. 

\section{Conclusion}\label{sec:conclusion}
The SRC algorithm classifies a test sample by seeking the sparsest representation in terms of the training data and does not work well under the open world assumption.  In this paper, we have introduced a training stage to the SRC algorithm so that it can be adapted to tackle the open set recognition problems.  The resulting algorithm makes use of the reconstruction error distributions modeled by the EVT.   Various experiments on popular image and object classification datasets have shown that our method can perform significantly better than many competitive open set recognition algorithms.

If the dataset contains extreme variations in pose, illumination or resolution, then the self expressiveness property required by the SRC algorithm will no longer hold.  In this case, the proposed SROSR algorithm will fail.  A possible solution to this problem would be to develop kernel-based methods for SROSR where kernel SRC \cite{SRC_MKL}, \cite{SR_kenrel}, \cite{Kernel_SRC} is used to find the sparse representation in the high-demential feature space.   Another limitation of the proposed SROSR method is that for good recognition performance, the training set  is  required  to  be  extensive  enough  to  span  the  conditions
that might occur in the test set.  Development of sparsity-based open set recognition method where only a single image or a very few images are given per class for training is an interesting open problem.  Furthermore, it remains an  interesting  topic  for  future work to develop a sparse representation or dictionary learning-based open set recognition algorithm by directly minimizing the open risk criteria.

\section*{Aknowledgement}
This work was supported by an ARO grant W911NF-16-1-0126. 

\bibliographystyle{IEEEtran}
\bibliography{openset}

\end{document}